\documentclass[letterpaper,10pt,conference]{ieeeconf}

\IEEEoverridecommandlockouts
\newif\ificrafinal
\icrafinaltrue

\usepackage{graphicx}
\usepackage{amsmath,amssymb,amsfonts}
\usepackage{bm}
\usepackage{booktabs}
\usepackage{multirow}
\usepackage{tabularx}
\usepackage{cite}
\usepackage{url}
\usepackage{xcolor}

\usepackage{algorithm}
\usepackage[noend]{algpseudocode}

\usepackage{acro}

\makeatletter
\let\NAT@parse\undefined
\makeatother
\usepackage[colorlinks=true,
            linkcolor=black,
            citecolor=black,
            urlcolor=blue]{hyperref}

\usepackage{latexml}
\usepackage{longtable}
\usepackage{epsfig}
\usepackage{epstopdf}
\usepackage{mathrsfs}
\usepackage{dsfont}
\usepackage{verbatim}
\usepackage{amscd}
\usepackage{rotating}
\usepackage[vcentermath]{youngtab}

\definecolor{lgray}{gray}{0.6}

\newcommand{\BoldB}				{ \mathbf{B} }

\newcommand{\BoldN}				{ \mathbf{N} }

\newcommand{\Boldp}				{ \mathbf{p} }
\newcommand{\BoldQ}				{ \mathbf{Q} }

\newcommand{\BoldR}				{ \mathbf{R} }

\newcommand{\Boldr}				{ \mathbf{r} }
\newcommand{\Bolds}				{ \mathbf{s} }

\newcommand{\BoldT}				{ \mathbf{T} }

\newcommand{\Boldv}				{ \mathbf{v} }

\newcommand{\Boldx}				{ \mathbf{x} }

\newcommand{\Boldz}				{ \mathbf{z} }

\newcommand{\1}					{ \boldsymbol{1} }

\newcommand{\Boldell}			{ \boldsymbol{\ell} }

\newcounter{inlineenum}
\renewcommand{\theinlineenum}{\alph{inlineenum}}

\DeclareMathOperator*{\argmin}{arg\,min}
\DeclareMathOperator*{\argmax}{arg\,max}

\newtheorem{Theorem}{Theorem}[section]

\newtheorem{Remark}[Theorem]{\textbf{Remark}}
\newcommand{\brmrk}[1]{\begin{Remark} \label{#1} }
\newcommand{\ermrk}{ \hfill $\bigtriangleup$    \end{Remark} \vspace{1mm} }

\newtheorem{exercise}{Exercise}[section]
\newtheorem{example}{Example}[section]
\newcommand{\boex}[1]{\begin{example} \label{#1} --- \rm}
\newcommand{\eoex}{ \hfill $\bigtriangleup$    \end{example} \vspace{1mm} }
\newcommand{\bohw}[1]{\begin{exercise} \label{#1} -- \rm}
\newcommand{\eohw}{ \hfill    \end{exercise} \vspace{1mm} }
\newtheorem{assumption}{Assumption}[section]
\newcommand{\boass}[1]{\begin{assumption} \label{#1} -- \rm}
\newcommand{\eoass}{ \hfill    \end{assumption} \vspace{1mm} }

\ificrafinal
	\newcommand{\todo}[1]{}
\else
	\newcommand{\todo}[1]{\footnote{\color{green}TO DO: {#1}}}
\fi

\renewcommand{\baselinestretch}{1.0}

\definecolor{brinkpink}{rgb}{1.00, 0.33, 0.64}



\DeclareAcronym{3DMA}{
	short=3DMA,
	long=3D Map Aiding,
}

\DeclareAcronym{APC}{
	short=APC,
	long=Antenna Phase Center,
}

\DeclareAcronym{COM}{
short=COM,
long=Center of Mass,
}

\DeclareAcronym{CV}{
short=CV,
long=Connected Vehicle,
}

\DeclareAcronym{CAV}{
	short=CAV,
	long=Connected Automated Vehicles,
}

\DeclareAcronym{IAR}{
short=IAR,
long=Integer Ambiguity Resolution,
}

\DeclareAcronym{DCS}{
	short=DCS,
	long=Dynamic Covariance Scaling,
}

\DeclareAcronym{CDMA}{
	short=CDMA,
	long=Code Division Multiple Access,
}

\DeclareAcronym{CDF}{
	short=CDF,
	long=Cumulative Distribution Function,
}

\DeclareAcronym{CE-CERT}{
	short=CE-CERT,
	long=College of Engineering Center for Environmental Research and Technology,
}

\DeclareAcronym{CME}{
	short=CME,
	long=common-mode errors,
}

\DeclareAcronym{ENU}{
	short=ENU,
	long=East-North-Up
}

\DeclareAcronym{FDMA}{
	short=FDMA,
	long=Frequency Division Multiple Access,
}

\DeclareAcronym{DD}{
	short=DD,
	long=Double-Differenced,
}

\DeclareAcronym{DGPS}{
	short=DGPS,
	long=Differential Global Positioning System,
}

\DeclareAcronym{DLR}{
	short=DLR,
	long=German Aerospace Center,
}
\DeclareAcronym{DGNSS}{
short=DGNSS,
long=Differential GNSS,
}
\DeclareAcronym{ECEF}{
short=ECEF,
long=Earth-Centered Earth-Fixed,
}
\DeclareAcronym{ECI}{
	short=ECI,
	long=Earth-Centered Inertial,
}
\DeclareAcronym{KF}{
	short=KF,
	long=Kalman Filter,
}
\DeclareAcronym{EKF}{
	short=EKF,
	long=Extended KF,
}

\DeclareAcronym{FGO}{
	short=FGO,
	long=Factor Graph Optimization,
}

\DeclareAcronym{GPS}{
	short=GPS,
	long=Global Positioning System,
}
\DeclareAcronym{GNSS}{
short=GNSS,
long=Global Navigation Satellite Systems,
}
\DeclareAcronym{GPSSPS}{
short=GPS SPS,
long=GPS standard positioning service,
}

\DeclareAcronym{HMM}
{short = HMM, long = Hidden Markov Model}

\DeclareAcronym{HD}
{short = HD, long = Horizontal Distance}

\DeclareAcronym{HiD}{
	short=Hi-Def,
	long=High-definition,
}

\DeclareAcronym{IMU}{
short=IMU,
long=Inertial Measurement Unit,
}

\DeclareAcronym{INS}{
	short=INS,
	long=Inertial Navigation System,
}

\DeclareAcronym{ILP}{
short=ILP,
long=Integer Linear Programming,
}

\DeclareAcronym{I2V}{
short=I2V,
long=Infrastructure-to-Vehicle,
}
\DeclareAcronym{IGS}{
	short=IGS,
	long=International GNSS Service,
}
\DeclareAcronym{ISB}{
	short=ISB,
	long=inter-system bias,
}

\DeclareAcronym{LLMM}
{short = LLMM, long = Lane-level Map-matching}

\DeclareAcronym{LoD}
{short = LoD, long = Level of Detail}

\DeclareAcronym{LD}
{short = LD, long = Lane Determination}

\DeclareAcronym{LOS}
{short = LOS, long = line-of-sight}

\DeclareAcronym{NLOS}
{short = NLOS, long = non-line-of-sight}

\DeclareAcronym{LMI}
{short = LMI, long = Linear Matrix Inequality}

\DeclareAcronym{RLMM}
{short = RLMM, long = Road-level Map-matching}

\DeclareAcronym{MSE}
{short = MSE, long = Mean Square Error}

\DeclareAcronym{MGEX}
{short = MGEX, long = Multi-GNSS Experiment}

\DeclareAcronym{MAP}
{short = MAP, long = Maximum-a-Posteriori}
\DeclareAcronym{OS}{
	short=OS,
	long=Open Service,
}

\DeclareAcronym{OSB}{
short=OSB,
long=Observable-specific Code Biases,
}
\DeclareAcronym{OSR}{
short=OSR,
long=Observation Space Representation,
}
\DeclareAcronym{PPP}{
	short=PPP,
	long=Precise Point Positioning,
} 

\DeclareAcronym{RTPPP}{
	short=RT-PPP,
	long=Real-time PPP,
}

\DeclareAcronym{PPP-AR}{
short=PPP-AR,
long=Precise Point Positioning Ambiguity Resolution,
} 

\DeclareAcronym{PP}{
short=PP,
long=Post Processing,
}

\DeclareAcronym{PVA}{
	short=PVA,
	long={position, velocity, acceleration},
} 

\DeclareAcronym{RTCM}{
short=RTCM,
long=Radio Technical Commission for Maritime Services,
}
\DeclareAcronym{RTK}{
short=RTK,
long=Real-time Kinematic,
}
\DeclareAcronym{SBAS}{
short=SBAS,
long=Satellite Based Augmentation Systems,
}

\DeclareAcronym{SNR}{
short=SNR,
long=Signal-to-Noise Ratio,
}
\DeclareAcronym{SSR}{
short=SSR,
long=State Space Representation,
}
\DeclareAcronym{SPS}{
	short=SPS,
	long=Standard Positioning Service,
}

\DeclareAcronym{SAE}{
	short=SAE,
	long=Society of Automotive Engineers,
}

\DeclareAcronym{STD}{
	short=STD,
	long=Standard Deviation,
}

\DeclareAcronym{STEC}{
	short=STEC,
	long=Slant Total Electron Content,
}
\DeclareAcronym{VTEC}{
	short=VTEC,
	long=Vertical Total Electron Content,
}

\DeclareAcronym{SH}{
	short=SH,
	long=Spherical Harmonic,
}
\DeclareAcronym{TGD}{
short=TGD,
long=Timing Group Delay,
}
\DeclareAcronym{TD}{
	short=TD,
	long=Threshold Decisions,
}
\DeclareAcronym{TOP}{
	short=TOP,
	long=time-of-signal-propagation,
}
\DeclareAcronym{TOT}{
	short=TOT,
	long=time-of-signal-transmission,
}
\DeclareAcronym{TOR}{
	short=TOR,
	long=time-of-signal-reception,
}
\DeclareAcronym{ZTD}{
short=ZTD,
long=Zenith Troposphere Delay,
}
\DeclareAcronym{TEC}{
short=TEC,
long=Total Electron Content,
}
\DeclareAcronym{IPP}{
	short=IPP,
	long=Ionosphere Pierce Point,
}
\DeclareAcronym{NOAA}{
	short=NOAA,
	long=National Oceanic and Atmospheric Administration,
}
\DeclareAcronym{UCR}{
	short=UCR,
	long=University of California-Riverside,
}
\DeclareAcronym{USTEC}{
short=US-TEC,
long=US Total Electron Content,
}
\DeclareAcronym{VNDGNSS}{
	short=VN-DGNSS,
	long=Virtual Network DGNSS,
}
\DeclareAcronym{IOD}{
	short=IOD,
	long=Issue Of Data,
}
\DeclareAcronym{CAS}{
	short=CAS,
	long=Chinese Academy of Sciences,
}
\DeclareAcronym{CNES}{
	short=CNES,
	long=Centre National d'Études Spatiales,
}
\DeclareAcronym{RMS}{
	short=RMS,
	long=Root Mean Square,
}
\DeclareAcronym{RW}{
	short=RW,
	long=random-walk,
}
\DeclareAcronym{RMSE}{
	short=RMSE,
	long=Root Mean Square Error,
}
\DeclareAcronym{SF}{
	short=SF,
	long=Single Frequency,
}

\DeclareAcronym{DF}{
	short=DF,
	long=Dual Frequency,
}
\DeclareAcronym{ICD}{
	short=ICD,
	long=Interface Control Document,
}
\DeclareAcronym{NED}{
	short=NED,
	long={North, East and Down},
}
\DeclareAcronym{WAAS}{
	short=WAAS,
	long=Wide Area Augmentation System,
}
\DeclareAcronym{OPUS}{
	short=OPUS,
	long=Online Positioning User Service,
}
\DeclareAcronym{PDF}{
	short=PDF,
	long=Probability Density Function,
}
\DeclareAcronym{RAPS}{
	short=RAPS,
	long=Risk-Averse Performance-Specified,
}

\DeclareAcronym{VRS}{
	short=VRS,
	long=Virtual Reference Station,
}

\DeclareAcronym{SPP}{
	short=SPP,
	long=Single-frequency Point Positioning,
}

\DeclareAcronym{SPaT}{
	short=SPaT,
	long=Signal Phase and Timing,
}

\DeclareAcronym{TECU}{
	short=TECU,
	long=Total Electron Content Units,
}

\DeclareAcronym{BNC}{
	short=BNC,
	long=BKG NTRIP Client,
}

\DeclareAcronym{ITS}{
	short=ITS,
	long=Intelligent Transportation Systems
}

\DeclareAcronym{USDOT}{
	short=USDOT,
	long=U.S. Department of Transportation
}

\DeclareAcronym{WHU}{
	short=WHU,
	long=Wuhan University
}

\DeclareAcronym{VINS}{
	short=VINS,
	long=Visual-Inertial Navigation System
}

\title{SeA-RVINS: Semantic-Aware Tightly Coupled RTK-Visual-Inertial System with Correlation-Preserving Robust Estimation for Urban Navigation}

\ificrafinal
  \iflatexml
  \author{Wang Hu$^{1}$ and Bo Wu$^{2}$
    \thanks{$^{1}$
      \texttt{huwang96@gmail.com};
      $^{2}$UC Riverside,
      \texttt{bwu109@ucr.edu}}%
  }
  \else
  \author{Wang Hu$^{1}$ and Bo Wu$^{2}$
    \thanks{$^{1}$
      \texttt{huwang96@gmail.com};
      $^{2}$UC Riverside,
      \texttt{bwu109@ucr.edu}}%
      \\
    {\normalfont\small
    \url{https://semantic-rvins.github.io/}}
  }
  \fi
\else
  \author{Anonymous Authors}
\fi

\begin{document}

\maketitle
\iflatexml
  \ificrafinal
    \begin{center}
      Project page: \url{https://semantic-rvins.github.io/}
    \end{center}
  \fi
\fi
\thispagestyle{empty}
\pagestyle{empty}

\begin{abstract}
  Reliable absolute pose estimation in urban environments is undermined by outlier measurements and incorrect
temporal associations that can persist in tightly coupled estimators.
Global Navigation Satellite System (GNSS) observations provide 
globally referenced measurements but are prone to multipath effects. Visual-inertial 
sensing supplies local motion constraints, but false visual 
associations can corrupt the estimator.
We present SeA-RVINS, a fixed-lag factor-graph Real-Time Kinematic (RTK)
visual-inertial system for robust urban pose estimation.
A semantic-aware learned stereo frontend rejects unreliable tracks before
persistent landmarks enter the graph.
For double-differenced GNSS measurements, SeA-RVINS applies Dynamic Covariance
Scaling through configurable batch, scalar, and latent-pivot robust formulations while retaining the
shared-pivot correlation structure.
We propose a hybrid ambiguity-continuation strategy that shares one ambiguity
state over short arcs with verified continuity and softly links successive arcs
through random-walk factors.
On an approximately 20-km route from the public TEX-CUP dataset, including about 50\% deep-urban driving, the
latent-pivot configuration achieves 100\% availability
and a 1.6-m maximum horizontal error, with 96.16\% and
99.90\% of epochs below 1.0 and 1.5~m, respectively.
The  implementation is released as open-source software\footnote{The source code will be released at GitHub upon acceptance.}.

\end{abstract}

\section{Introduction}
\label{sec:introduction}

Continuous and accurate absolute
pose estimation is essential for 
connected and automated vehicles and outdoor robots operating over large areas.
Beyond average accuracy, these systems require high availability and resistance to occasional large errors;
many connected-vehicle applications require or benefit from sub-meter horizontal accuracy \cite{9043735,hu2024square,wang2023give}.
Reliable localization is often achieved through fusion of \ac{GNSS}, particularly carrier-phase-based \ac{RTK}, with vision and an \ac{IMU}\cite{hu2024square,wang2023give,huang2024rtkvi,duan2025slidingambiguity},
but their failure modes interact in urban canyons:
\ac{GNSS} signals are degraded by blockage, multipath, and \ac{NLOS} reception; inertial errors accumulate over time; and visual tracks can be corrupted
by moving objects, repeated structures, weak texture, and poor stereo geometry 
\cite{georgiadou1988carrier,adham2025semanticgvins,li2025high,wang2026semanticstereo}.
Hereafter, multipath- and \ac{NLOS}-affected observations are referred to as \ac{GNSS} outliers.

RTK-Visual-Inertial estimators are commonly
implemented with sequential filtering or fixed-lag \ac{FGO}.
Tightly coupling raw measurements in \ac{FGO} improves information use, but
once a corrupted \ac{GNSS} factor, a false visual landmark, or a wrongly continued carrier-phase ambiguity enters the estimator, its influence can persist through states and marginalization. Reliable pose estimation therefore
requires more than sensor aggregation: outliers and incorrect
temporal associations must be rejected or appropriately downweighted to limit their influence.

Carrier-phase observations have millimeter-level noise and typically centimeter-level multipath, but each contains an unknown integer ambiguity~\cite{georgiadou1988carrier}. \ac{RTK} typically forms \ac{DD} measurements to mitigate common-mode errors and eliminate receiver clock bias~\cite{hu2025optimization}.
In the \ac{RTK} approach, \ac{IAR} is performed on the float \ac{DD} ambiguities.
The ambiguity remains constant but changes after a cycle slip or loss of lock~\cite{teunissen2017springer}. Reusing one ambiguity state across epochs strengthens float estimation and \ac{IAR}, whereas extending it across an undetected discontinuity can corrupt the estimator. Conversely, independent per-epoch ambiguities prevent such propagation but discard temporal information and enlarge the graph. Moreover, geometry-free cycle-slip tests require observations on at least two frequencies; when only single-frequency is available, continuity may be unverifiable in real time \cite{blewitt1990automatic}.
The \ac{DD} operation induces shared-pivot correlations within each batch
of measurements. Robustification should therefore limit outlier influence
while accounting for the correlation introduced by differencing
~\cite{wang2022outliercorrelated}.

Vision can aid urban navigation either through pose-level constraints or
through feature-level measurements. Pose-level aiding depends strongly on
the quality of the surviving static inliers,
while feature-level fusion can turn a false association into a persistent landmark that biases the estimates.
Moving objects, repeated facades and lane markings, weak texture, and poor stereo geometry make both cases difficult in urban driving \cite{adham2025semanticgvins,li2025high,wang2026semanticstereo}.
A reliable frontend must consequently do more than maximize the number of
matches: it should reject semantically unsafe regions, reduce ambiguous
matches, and prevent low-quality tracks from entering the estimator.


To address these challenges, we present an open-source SeA-RVINS,
a semantic-aware \ac{RTK} \ac{VINS} with the following main contributions:

\begin{itemize}
    \item A semantic-aware learned stereo frontend that rejects semantically unreliable and geometrically inconsistent tracks before they become persistent landmark factors.
    

    \item A hybrid ambiguity-continuation topology that shares states over bounded, continuity-verified arcs, softly links successive arcs, and initializes independent states when continuity is broken or unverifiable.
    

    \item A configurable correlation-preserving robust formulation with
    \ac{DCS}~\cite{agarwal2013robust} for \ac{DD} code and phase.
    A conventional batch mode provides a full-covariance reference;
    scalar reweighting independently weights the full-covariance-whitened
    components without adding graph variables, while the latent-pivot
    formulation introduces shared pivot-error variables to enable
    candidate-wise robust weighting.
    

    \item A full-route evaluation on the approximately 20-km TEX-CUP
    dataset, including about 50\% deep-urban driving and
    comparisons with publicly available SOTA sources.
    SeA-RVINS latent-pivot configuration achieves 100\% solution availability, a maximum horizontal error of 1.6~m, and 99.9\% of epochs below 1.5~m.
\end{itemize}

\section{Related Work}
\label{sec:related-work}

\subsection{\ac{RTK}-\ac{VINS} Estimation and Ambiguity Management}

GNSS-VINS integration may loosely couple processed pose estimates, or tightly couple raw sensor measurements.
GVINS~\cite{cao2022gvins} tightly couples GNSS measurements with visual--inertial states;
GICI-LIB~\cite{chi2023gici} provides open \ac{FGO} implementations of several raw-GNSS modes.
Wen et al.~\cite{wen20233d} integrate \ac{RTK} with forward-camera landmarks and a sky-pointing camera for \ac{NLOS} rejection; filter-based alternatives include SRI-GVINS~\cite{hu2024square},
which fuses raw measurements in a square-root inverse sliding-window filter.
The temporal treatment of carrier-phase ambiguities is a central issue in \ac{RTK} integration.
GIVE~\cite{wang2023give} propagates continuously tracked ambiguity information.
Huang et al.~\cite{huang2024rtkvi} retain \ac{GNSS} states to recover the covariance required for ambiguity resolution, and
Duan et al.\cite{duan2025slidingambiguity} maintain a sliding ambiguity window with visual--inertial-aided cycle-slip detection.
These works establish the value of temporal ambiguity reuse. Our work addresses the complementary case in which continuity is only partially observable: exact state sharing is bounded, successive bounded segments are softly linked, and unverifiable continuity starts an independent ambiguity state.

\subsection{Reliable Visual Frontends in Urban Environments}

Most raw-\ac{GNSS} VINS retain conventional sparse frontends based on corners, optical flow, or local descriptors;
for example, GVINS uses a monocular feature-based visual frontend, while GICI-LIB and
SRI-GVINS employ conventional visual processing and mainly advance GNSS integration and estimator design.
Such frontends can perform well in favorable scenes, but
local texture alone does not indicate whether a feature lies
on a moving object, and repeated urban structures may
produce several locally similar matching candidates.
Recent work has begun to address these limitations.
Adham et al.~\cite{adham2025semanticgvins} combine semantic feature grading and
descriptor matching to suppress dynamic features.
Wang et al.~\cite{wang2026semanticstereo} apply semantic segmentation and stereo epipolar verification, but use standard GNSS positioning rather than raw carrier-phase measurements.
Li et al.~\cite{li2025high} combine YOLO~\cite{wang2024yolov10} for dynamic-region detection, optical-flow tracking, and an RTK stochastic model within an \ac{FGO} framework.
These approaches improve visual robustness primarily through semantic and geometric
filtering of dynamic features, but do not combine learned feature extraction with
learned stereo and temporal matching to reduce ambiguous static associations.
SeA-RVINS instead combines SegFormer~\cite{xie2021segformer} semantic screening of both rectified stereo views, SuperPoint--LightGlue~\cite{detone2018superpoint,lindenberger2023lightglue} feature extraction and association across stereo and temporal views, and track- and frame-level motion checks before visual factors enter the graph. The contribution is therefore
not semantics alone, but
a semantic-aware learned visual factor-admission pipeline for persistent landmarks in an \ac{RTK} \ac{FGO} estimator.


\subsection{Correlation-Preserving Robust GNSS Factors}



Conventional residual tests may not reject all \ac{GNSS} outliers, motivating robust estimation in the estimator~\cite{wen2022gnc,hu2025optimization}.
\ac{DCS} is attractive for \ac{FGO} because it retains nominal information for small residuals and downweights
large ones without introducing a switch variable for every factor~\cite{agarwal2013robust}.
For \ac{DD} measurements, candidates sharing a pivot have correlated
errors. A standard batch formulation uses the full covariance and applies one robust loss to the batch Mahalanobis norm. This accounts for the correlation but assigns one robust weight to the entire batch, so an outlier can downweight otherwise clean observations.
Existing \ac{RTK} estimators preserve this dependence through pre-differenced states or explicit covariance handling~\cite{chi2023gici,huang2024rtkvi}, consistent with the Gaussian
equivalence between estimating shared bias terms and using weighted differenced observations~\cite{blewitt1997basics,platz2023equivalence}.
To our knowledge, prior RTK--visual--inertial formulations have not enabled candidate-wise robustification while preserving the shared-pivot correlation structure of \ac{DD} measurements.
Hereafter, \emph{batch-wise} weighting denotes one robust weight
for the correlated residual vector; \emph{component-wise} weighting
assigns separate weights after full-covariance whitening; and
\emph{candidate-wise} weighting assigns separate weights to
individual candidate--pivot residuals.


SeA-RVINS implements and compares these three configurations for DD code and phase with \ac{DCS} robust weighting in a common estimator as described in Sec.~\ref{subsec:robust_est}.
All formulations recover the
same full-covariance Gaussian model under unit robust weights.

\section{Proposed Method}
\label{sec:method}

\subsection{System Overview}
\label{subsec:system-overview}

\begin{figure*}[t]
    \centering
    \includegraphics[
        width=0.87\textwidth,
        keepaspectratio
    ]{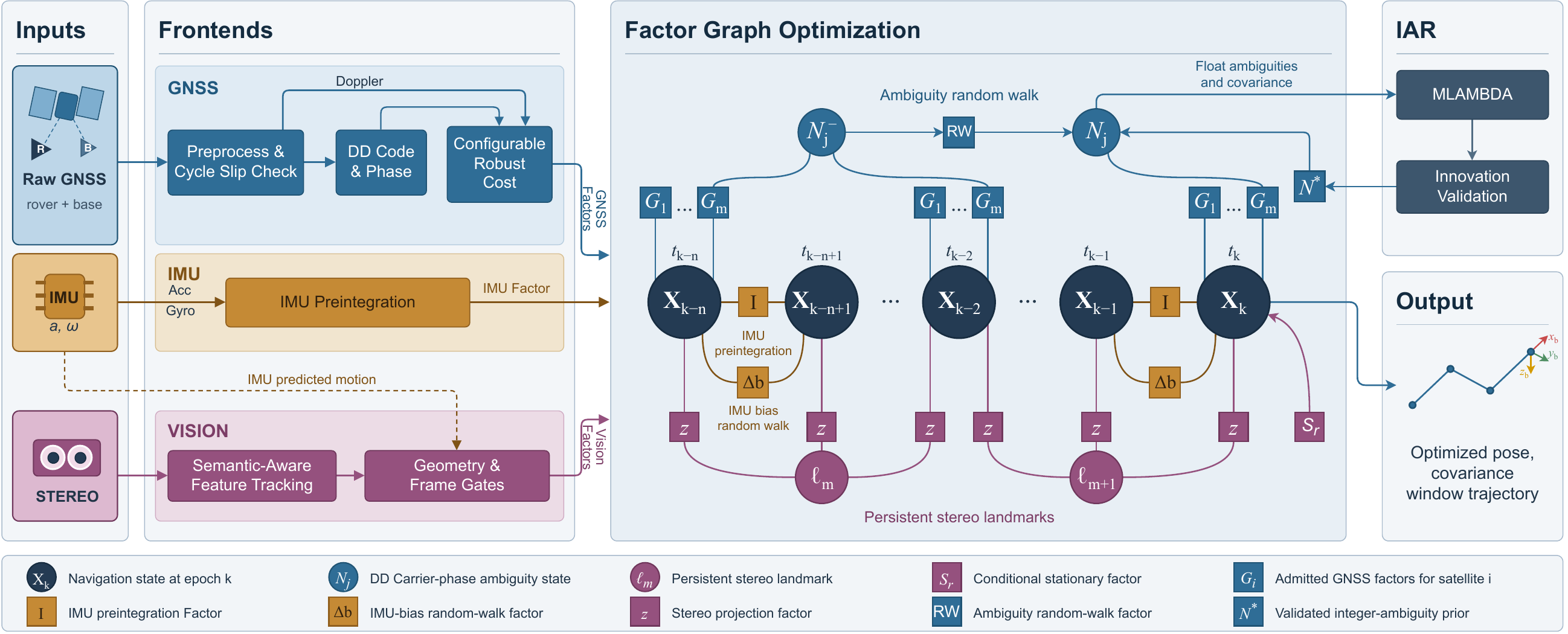}
    \caption{System overview and schematic \ac{FGO}.
Latent pivot variables and marginalization are
omitted for clarity. The stereo frontend is detailed in
Fig.~\ref{fig:vision_frontend}.}
    \label{fig:system_overview}
\end{figure*}

Fig.~\ref{fig:system_overview} illustrates the proposed SeA-RVINS, which tightly couples GNSS, IMU, and stereo observations in an \ac{FGO} framework.
\ac{IMU} preintegration links successive navigation states, while the semantic-aware stereo frontend supplies persistent landmark factors and conditional stationary cues.
Rover and base \ac{GNSS} observations undergo preprocessing, cycle-slip detection, and pivot selection before \ac{DD} code and phase factors are constructed; Doppler observations provide velocity constraints.
The graph jointly estimates navigation states,
landmarks, float ambiguities, and optional pivot-error and
stationary-anchor variables.
Configurable correlation-preserving robust formulations are applied to differenced \ac{GNSS} measurements, as described in Sec.~\ref{subsec:robust_est}.
Ambiguity continuation is detailed in Sec.~\ref{subsec:ambiguity-management}.
\ac{IAR} uses the float ambiguities and their joint covariance, and only validated integer information is returned to the graph. The optimized states provide the current 6-DoF pose and window trajectory.

Let $w$, $b$, $c_l$, and $e$ denote the local \ac{ENU} world frame, \ac{IMU} body
frame, rectified left-camera frame, and \ac{ECEF} frame, respectively. The navigation state at an epoch $k$ is
\begin{equation}
    \mathbf{x}_k \triangleq
    \left(
        \mathbf{R}^{w}_{b_k},
        {}^{w}\mathbf{p}_{b_k},
        {}^{w}\mathbf{v}_{b_k},
        \BoldB_{a, k},
        \BoldB_{g, k}
    \right),
    \label{eq:navigation-state}
\end{equation}
where $\mathbf{R}^{w}_{b_k}\in\mathrm{SO}(3)$, and $\BoldB_{a, k}$ and
$\BoldB_{g, k}$ are the accelerometer and gyroscope biases. For a
fixed-lag window $\mathcal{W}$, the complete variable set is
\begin{equation}
\begin{split}
    \mathcal{X}_{\mathcal{W}} \triangleq {}&
    \{\mathbf{x}_k\}_{k\in\mathcal{W}}
    \cup \{{}^{w}\boldsymbol{\ell}_j\}_{j\in\mathcal{L}}
    \cup \{N_{\kappa}\}_{\kappa\in\mathcal{A}} \\
    &\cup \{n^t_{k,q}\}_{(k,q)\in\mathcal{G}_t,\,t\in\{\rho,\Phi,D\}}
    \cup \{{}^{w}\mathbf{s}_r\},
    \label{eq:window-state}
\end{split}
\end{equation}
where ${}^{w}\boldsymbol{\ell}_j$ is a persistent stereo landmark,
$N_{\kappa}$ is a \ac{DD} ambiguity associated with a graph variable node $\kappa$,
$\mathcal{L}$ and $\mathcal{A}$ are the active landmark and ambiguity index sets,
$n^t_{k,q}$ is the latent pivot error variable for \ac{GNSS} batch $q$ for observation type $t$, and ${}^{w}\mathbf{s}_r$ is an optional stationary anchor.
The set $\mathcal G_t$ contains batches represented by latent pivot-error variables and is empty under the batch and scalar configurations.
The antenna position and velocity are defined as
\begin{align}
    {}^{e}\mathbf{p}_{ant, k}(\Boldx)
    &={}^{e}\mathbf{p}_{0}
    +\mathbf{R}^{e}_{w}
    \left({}^{w}\mathbf{p}_{b_k}
    +\mathbf{R}^{w}_{b_k}\,{}^{b}\mathbf{r}_{ant}\right),
    \label{eq:antenna-position} \\
    {}^e\Boldv_{ant, k}(\Boldx)
    &=\mathbf{R}^e_w\!\left[
      {}^w\Boldv_{b_k}+
      \mathbf{R}^w_{b_k}\!\left(
        \widehat{\boldsymbol{\omega}}_{eb}^{b}
        \times{}^b\Boldr_{ant}\right)\right],
    \label{eq:antenna-velocity}
\end{align}
where ${}^{e}\mathbf{p}_{0}$ is the \ac{ECEF} origin of the local frame, $\widehat{\boldsymbol{\omega}}_{eb}^{b}$ is the body angular rate relative to the Earth, expressed in frame $b$ and ${}^{b}\mathbf{r}_{ant}$ is the known \ac{IMU}-to-antenna lever arm.

Let $\mathcal{Z}_{\mathcal{W}}$ denote the measurements associated with the
current fixed-lag window. 
The \ac{FGO} represents the factorized posterior
$p(\mathcal{X}_{\mathcal{W}}|\mathcal{Z}_{\mathcal{W}})$, and the
\ac{MAP} estimate is obtained by minimizing its negative logarithm:
\begin{equation}
\begin{aligned}
    \mathcal{X}_{\mathcal{W}}^{\star}
    &= \argmax_{\mathcal{X}_{\mathcal{W}}}
       p(\mathcal{X}_{\mathcal{W}}\mid\mathcal{Z}_{\mathcal{W}})\\
    &= \argmin_{\mathcal{X}_{\mathcal{W}}}\big\{
       \mathcal{J}_{\mathrm{prior}}+\mathcal{J}_{I}+\mathcal{J}_{B}
       +\mathcal{J}_{V}\\[-1mm]
    &\hspace{26mm}
       +\mathcal{J}_{\mathrm{DD}}+\mathcal{J}_{D}
       +\mathcal{J}_{N}+\mathcal{J}_{S}\big\}.
    \label{eq:factor-graph-objective}
\end{aligned}
\end{equation}
where $\mathcal{J}_{\mathrm{prior}}$, $\mathcal{J}_{I}$, $\mathcal{J}_{B}$,
$\mathcal{J}_{V}$, $\mathcal{J}_{\mathrm{DD}}$, $\mathcal{J}_{D}$,
$\mathcal{J}_{N}$, and $\mathcal{J}_{S}$ denote the marginalized
prior, IMU-preintegration, bias \ac{RW}, stereo-reprojection, \ac{DD}
code/phase, Doppler, ambiguity, and stationary costs, respectively. Each
$\mathcal{J}_{\cdot}$ collects the residual costs of the corresponding measurement factors.
We use $\|\Boldr\|_{\boldsymbol{\Sigma}}^2
\triangleq\Boldr^{\mathsf T}\boldsymbol{\Sigma}^{-1}\Boldr$.
Gaussian factors contribute half of the squared Mahalanobis norm.
Stereo factors use the Huber loss~\cite{huber1964robust}.
The robust \ac{GNSS} costs $\mathcal{J}_{\mathrm{DD}}$ and
$\mathcal{J}_{D}$ using correlation-preserving
representation are defined in Sec.~\ref{subsec:robust_est}.
Variables leaving the fixed-lag window are
marginalized, and their information is retained in
$\mathcal{J}_{\mathrm{prior}}$.


\subsection{\ac{IMU} Preintegration}
\label{subsec:imu-preintegration}

The \ac{IMU} provides high-rate angular-velocity and specific-force measurements between lower-rate graph states associated with \ac{GNSS} epochs and camera epochs.
The standard on-manifold
preintegration formulation of \cite{forster2015imu} is used to summarize \ac{IMU}
samples between consecutive graph epochs as a single relative-motion constraint.

For consecutive epochs $i$ and $j$, the intervening measurements are
summarized by
$\Delta\widehat{\Boldz}^{I}_{ij} \triangleq \left(\Delta\widehat{\BoldR}_{ij}, \Delta\widehat{\Boldv}_{ij}, \Delta\widehat{\Boldp}_{ij}\right)$
where $\Delta \hat{\mathbf R}_{ij}$,
$\Delta \hat{\mathbf v}_{ij}$, and
$\Delta \hat{\mathbf p}_{ij}$ are the preintegrated rotation, velocity, and
position increments expressed in frame $b_i$, respectively, and
$\Sigma^{I}_{ij}$ is their propagated covariance.

The bias-corrected preintegration residual
$\Boldr^{I}_{ij}$ follows Eqns.~(36)--(37) in \cite{forster2015imu}.
Accordingly, the \ac{IMU} preintegration and bias \ac{RW} terms
in eqn.~\eqref{eq:factor-graph-objective} are
\begin{align}
\mathcal{J}_I =
\frac{1}{2}
\sum_{(i,j)\in\mathcal{E}_I}
\left\|
    \Boldr^{I}_{ij}
\right\|_{\Sigma^{I}_{ij}}^{2},
\,
\mathcal{J}_B =
\frac{1}{2}
\sum_{(i,j)\in\mathcal{E}_I}
\left\|
    \BoldB_j-\BoldB_i
\right\|_{\Sigma^{B}_{ij}}^{2},
\end{align}
where $\mathcal{E}_I$ denotes the set of consecutive graph-epoch pairs,
$\BoldB_k
\triangleq
[\BoldB_{a,k}^{\top},\BoldB_{g,k}^{\top}]^{\top}$,
and $\Sigma^{B}_{ij}$ is the bias \ac{RW} covariance.


\subsection{Semantic-Aware Stereo Vision Frontend}
\label{subsec:semantic-vision}

\begin{figure}[t]
    \centering
    \includegraphics[width=1.0\columnwidth]{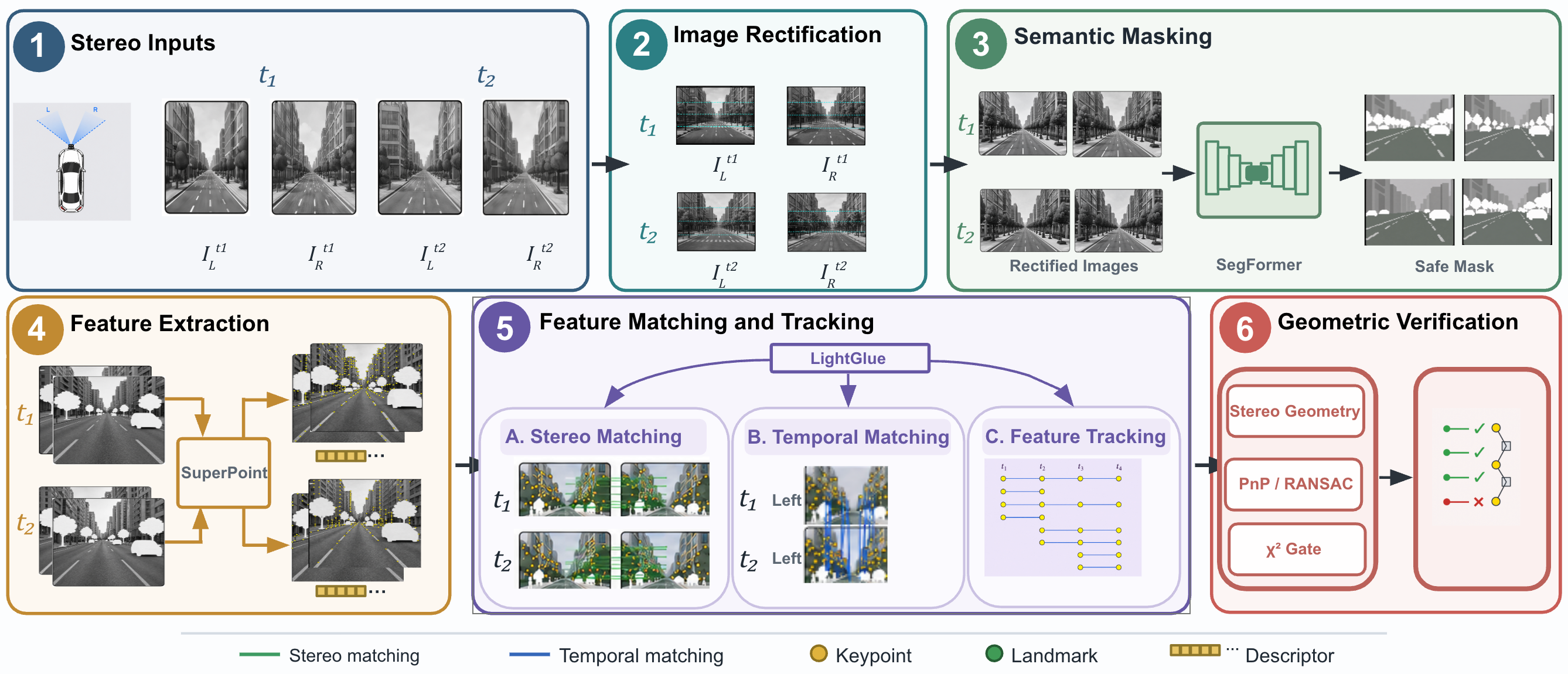}
    \caption{Workflow of the semantic-aware learned stereo frontend for feature
    tracking and landmark admission.}
    \label{fig:vision_frontend}
\end{figure}

Urban driving scenes contain moving traffic, repeated structures, and regions
with weak stereo geometry. The objective of the proposed frontend is consequently not
to maximize the number of matches, but to deliver semantically screened and geometrically consistent stereo observations to the estimator.
Fig.~\ref{fig:vision_frontend} expands the vision branch of the system overview in Fig.~\ref{fig:system_overview}, which runs in a dedicated thread.

At camera epoch $k$, synchronized grayscale stereo images $I^L_k$ and $I^R_k$ are rectified.
SegFormer~\cite{xie2021segformer} independently segments both views
to mask semantically unreliable classes, including sky, vegetation,
people, and vehicles. Mask erosion and additional buffers around
rejected regions provide conservative object-boundary margins.
SuperPoint jointly detects keypoints and computes their
descriptors~\cite{detone2018superpoint}; only keypoints satisfying the
corresponding left or right mask are retained.

LightGlue~\cite{lindenberger2023lightglue} performs both stereo and temporal
matching. Left--right matching within the same epoch provides stereo correspondences,
whereas matching $I^L_{k-1}$ to $I^L_k$ establishes temporal tracking.
Stereo matches undergo match-score and geometric checks, including epipolar,
vertical-disparity, and positive-disparity tests, before
triangulation~\cite{hartley2004multiple}. Triangulated points with invalid
geometry or excessive range are subsequently removed.
For the temporal branch, the previous stereo points and their current left-image
observations are passed to the PnP/RANSAC consistency check~\cite{fischler1981random}.
Its outputs are an inlier set, an auxiliary motion estimate
$\Delta\widehat{\BoldT}^{\mathrm{PnP}}_{k-1,k}$, and reprojection statistics.
Rejected associations break their tracks, which must mature again before landmark initialization.
The PnP support statistics scale the stereo-factor covariance. Separately, consistency between
$\Delta\widehat{\BoldT}^{\mathrm{PnP}}_{k-1,k}$ and the IMU-predicted motion over the same interval controls visual-factor admission, which determines whether the \ac{FGO}
initializes new landmarks, reuses existing landmarks, or rejects the frame.
The auxiliary PnP motion also supports stationary detection.


A track becomes eligible for landmark initialization after at least
$n_{\min}$ observations. For an admitted observation
$\Boldz^V_{kj}=[u^L_{kj},u^R_{kj},v_{kj}]^{\mathsf T}$,
the factor connects body pose $\BoldT^w_{b_k}$ and landmark
${}^w\Boldell_j$ through
\begin{equation}
\begin{aligned}
    \Boldr^V_{kj}
    &=\Boldz^V_{kj}-\pi_s\!\left[
      (\BoldT^w_{b_k}\BoldT^b_{c_l})^{-1}\cdot{}^w\Boldell_j
    \right],\\
    \mathcal{J}_V
    &=\sum_{(k,j)\in\mathcal{E}_V}
      \mathcal{L}_{\mathrm H}\!\left(
        \|\Boldr^V_{kj}\|_{\boldsymbol{\Sigma}^V_{kj}};
        c_V\right).
\end{aligned}
\label{eq:stereo-factor}
\end{equation}
where $\pi_s$ is the calibrated rectified-stereo projection into
$[u^L,u^R,v]^{\mathsf T}$~\cite{hartley2004multiple}, $\Boldr^V_{kj}$ is the stereo reprojection residual of landmark $j$,
$\BoldT^b_{c_l}$ maps the rectified left-camera frame to the body
frame, $\mathcal{E}_V$ is the set of admitted stereo
observations in the active window, and
$\boldsymbol{\Sigma}^V_{kj}$ is the covariance of the stereo reprojection residual.
The Huber loss $\mathcal{L}_{\mathrm H}(z;c_V)$ has threshold $c_V$ and quadratic branch $z^2/2$~\cite{huber1964robust}.
Multiple admitted observations share the same landmark variable.

When PnP-derived vision motion indicates a stop,
the backend can add a zero-velocity (ZUPT)
constraint and, under weak GNSS support, tie stopped poses to a
shared position anchor ${}^w\Bolds_r$. The residuals are
\begin{equation}
    \Boldr^{\mathrm{ZUPT}}_k={}^w\Boldv_{b_k},
    \qquad
    \Boldr^{\mathrm{anchor}}_{k,r}
    ={}^w\Boldp_{b_k}-{}^w\Bolds_r.
    \label{eq:vision-stationary-factors}
\end{equation}
The anchor carries a covariance-derived prior from a latched
stopped-position estimate. These Gaussian constraints and the
anchor prior contribute to $\mathcal{J}_S$.

\subsection{\ac{GNSS} Measurements}
\label{subsec:gnss-measurements}

For the short-baseline \ac{RTK} considered here,
the \ac{DD} operation performs rover--base differencing to mitigate common-mode errors, and candidate--pivot differencing within each constellation and signal type to eliminate common receiver clock and hardware biases~\cite{hu2025optimization,farrell2008aided}.
The pivot satellite is selected using high elevation and $C/N_0$ with lower multipath risk.
The base-to-satellite range terms are computed from the known base position and compensated during \ac{DD} formation.

For candidate satellite $s$ and pivot satellite $o$ on frequency $f$, omitting epoch index $k$ in subsequent equations,
the \ac{DD} code and phase measurement models are  (Sec. 8.8.3 in~\cite{farrell2008aided})
\begin{align}
	\rho^{s,o}_{f} &= h^{s,o}(\Boldx) + M_{\rho, f}^{s,o} + \eta^{s,o}_{\rho, f},\\
	\lambda_f \phi^{s,o}_{f} &= h^{s,o}(\Boldx) + \lambda_f N^{s,o}_f + M_{\phi, f}^{s,o} + \eta^{s,o}_{\phi, f}
    \label{eq:dd-code-phase-model}
\end{align}
where $h^{s,o}(\Boldx)\triangleq h^s(\Boldx)-h^o(\Boldx)$ with $h^s(\Boldx)=\|{}^e\Boldp_{ant}-{}^e\Boldp_s\|+\delta\rho^s_{\mathrm{Sag}}$ defines the geometric range where ${}^{e}\mathbf{p}_{s}$ is the satellite position and $\delta\rho^{s}_{\mathrm{Sag}}$ is the Sagnac correction \cite{hu2024sagnac}, ${}^e\Boldp_{ant}$ follows eqn.~\eqref{eq:antenna-position}, $\lambda_f$ is the wavelength,
$M^{s,o} = M^s-M^o$ represents the \ac{DD} multipath and residual errors, $N^{s,o} = N^s-N^o$ is the pivot-relative \ac{DD} integer ambiguity, discussed in
Sec.~\ref{subsec:ambiguity-management}, and $\eta^{s,o} = \eta^s - \eta^o$ is the measurement noise term. The combined multipath and measurement noise of satellite $i$ is assumed to be zero-mean Gaussian, represented as $(M_{f}^{i} + \eta^{i}_{f}) \sim \mathcal{N}(0, \sigma^2_i)$, where $\sigma^2_i$ is an elevation-dependent variance. Outliers represent departures from this Gaussian model.
The corresponding residuals are
\begin{align}
    r^{\rho}_{s,f}
    &=\rho^{s,o}_f-h^{s,o}(\Boldx), \label{eq:dd-code-residual}\\
    r^{\Phi}_{s,f}
    &=\lambda_f\phi^{s,o}_f-h^{s,o}(\Boldx)
      -\lambda_f N^{s,o}_f.
\label{eq:dd-phase-residual}
\end{align}

After Doppler observations are converted to range rate and
compensated for satellite velocity and clock drift,
between-satellite differencing removes receiver clock drift.
Let
$\1^s=({}^e\Boldp_{ant}-{}^e\Boldp_s)/
\|{}^e\Boldp_{ant}-{}^e\Boldp_s\|$
be the \ac{LOS} unit vector. The corrected Doppler model and its residual are
\begin{align}
    \delta D^{s,o}_f &= (\1^s - \1^o)^\top \, {}^{e}\Boldv_{ant} (\Boldx) + \eta^{s,o}_{D,f}, \label{eqn:linear_dop} \\
    r^D_{s,f}&=\delta D^{s,o}_f
      -(\1^s-\1^o)^{\mathsf T}{}^e\Boldv_{ant}(\Boldx).
    \label{eq:dd-doppler-residual}
\end{align}
where $\Boldv_{ant}$ follows eqn.~\eqref{eq:antenna-velocity}, $\eta^{s,o}_{D,f} =(\eta^s_{D,f}-\eta^o_{D,f})$, and $\eta^i_{D,f} \sim \mathcal{N}(0, \sigma^2_{i,D})$ is the measurement noise.

Because differencing creates correlations, the residuals sharing pivot $o$ have covariance
\begin{equation}
    [\boldsymbol{\Sigma}^{t}_{q}]_{ij}
    =\mathbb{I}[i=j]\sigma^2_{t,i}+\sigma^2_{t,o},
    \quad t\in\{\rho,\Phi,D\},
    \label{eq:dd-covariance}
\end{equation}
where $q$ denotes a batch of
$m$ candidate--pivot residuals sharing a pivot observation $o$ and $\mathbb{I}$ is the indicator function.
The correlation-preserving robust formulations for code, carrier phase,
and Doppler are described in Sec.~\ref{subsec:robust_est}.

\subsection{Correlation-Preserving Robust Estimation}
\label{subsec:robust_est}

Robust M-estimation replaces the quadratic residual cost $z^2/2$
with a robust loss $\mathcal L_{\mathrm{rob}}(z)$ to limit the
influence of outliers~\cite{huber2robust}.
Let $\mathbf r\in\mathbb R^m$ be a residual batch with covariance
$\boldsymbol\Sigma$, and let $\mathbf e=\mathbf L\mathbf r$ be its
whitened form, where
$\mathbf L^{\mathsf T}\mathbf L=\boldsymbol\Sigma^{-1}$.
SeA-RVINS supports three configurable modes:

\noindent\textit{\textbf{1. Batch formulation:}}
The conventional batch formulation retains the full covariance and
applies one robust weight to the entire whitened residual vector:
\begin{equation}
    \widetilde{\mathcal J}^{t}_{q,\mathrm B}
    =
    \frac{1}{2}w^t_{q,\mathrm B}
    \|\mathbf e^t\|_2^2 .
    \label{eq:batch-robust-objective}
\end{equation}
It requires no additional graph variables, but one outlier can
downweight otherwise clean measurements in the same batch.
The scalar and latent-pivot formulations below provide finer-grained
robust weighting while retaining the nominal shared-pivot dependence.

\noindent\textit{\textbf{2. Scalar formulation:}}
The scalar formulation retains the full covariance and applies a separate robust weight to each
whitened component. With the weights held fixed, its local
reweighted least-squares cost is
\begin{equation}
    \widetilde{\mathcal J}^{t}_{q,\mathrm S}
    =
    \frac{1}{2}
    (\mathbf e^t)^{\mathsf T}
    \mathbf W_q^t
    \mathbf e^t,
    \quad
    \mathbf W_q^t
    =
    \operatorname{diag}(w_1^t,\ldots,w_m^t).
    \label{eq:scalar-robust-objective}
\end{equation}
This formulation requires no additional graph variables and retains the correlation
through the whitened form.
However, because $\mathbf L_q^t$ is generally dense, each whitened
component can contain contributions from multiple original candidate
residuals; an outlier can therefore affect component
weights~\cite{wang2022outliercorrelated}.

\noindent\textit{\textbf{3. Latent-pivot formulation:}}
The shared-pivot covariance in eqn.~\eqref{eq:dd-covariance} can be
written as a matrix
\begin{equation}
    \boldsymbol\Sigma_q^t
    =
    \mathbf D^t
    +
    \sigma^2_{t,o}
    \boldsymbol\ell^t(\boldsymbol\ell^t)^{\mathsf T},
    \label{eq:latent-pivot-decomposition}
\end{equation}
where
$\mathbf D^t=\operatorname{diag}(d_1^t,\ldots,d_m^t)$ contains
the candidate variances excluding the pivot contribution and
$\ell_i^t$ is the pivot-error coefficient.
For eqn.~\eqref{eq:dd-covariance},
$d_i^t=\sigma^2_{t,i}$ and $\ell_i^t=1$.

The latent-pivot formulation introduces one epoch-local Gaussian
pivot variable for each batch and constructs the standardized scalar residual:
\begin{equation}
    n_q^t\sim\mathcal N(0,\sigma^2_{t,o}),
    \quad
    u_i^t=(r_i^t+\ell_i^t n_q^t) / \sqrt{d_i^t}.
    \label{eq:latent-pivot-residual}
\end{equation}
where $r_i^t$ denotes its residual in eqns.~\eqref{eq:dd-code-residual}, ~\eqref{eq:dd-phase-residual}
or~\eqref{eq:dd-doppler-residual}, evaluated at the current optimization iterate.
Marginalizing $n_q^t$ recovers eqn.~\eqref{eq:latent-pivot-decomposition},
whereas, under the assumed Gaussian model, conditioning on it yields independent scalar residuals tied to the
original candidates.
Code, carrier phase, and Doppler use separate epoch-local pivot-error variables. 
With robust weights held fixed, the local reweighted least-squares cost is
\begin{equation}
    \widetilde{\mathcal J}^{t}_{q,\mathrm L}
    =
    \frac{(n_q^t)^2}{2\sigma^2_{t,o}}
    +
    \frac{1}{2}\sum_{i=1}^{m}w_i^t(u_i^t)^2 .
    \label{eq:latent-pivot-objective}
\end{equation}

\noindent\textit{\textbf{DCS robust weighting:}}
For code and phase, let $z$ denote the normalized residual
quantity to which \ac{DCS} is applied. The corresponding
information-weight function~\cite{agarwal2013robust} is
\begin{equation}
    w_{\mathrm{DCS}}(z)
    =
    \left[
        \min\!\left(
            1,
            \frac{2c_{\mathrm{DCS}}}
                 {c_{\mathrm{DCS}}+z^2}
        \right)
    \right]^2 ,
    \label{eq:dcs-weight}
\end{equation}
where $c_{\mathrm{DCS}}>0$.
For the batch, scalar, and latent-pivot configurations,
$z=\|\mathbf e^t\|_2$, $z=e_i^t$, and $z=u_i^t$,
respectively.
Thus, the three configurations assign one weight per batch,
per whitened component, or per candidate residual, respectively.
Doppler uses Huber weighting with the corresponding robust
argument under the same selected configuration.
Weights are recomputed at each relinearization and held fixed during
the local least-squares solve. Latent pivot-error priors remain
Gaussian.
Summing the selected local costs over batches and epochs gives the
reweighted forms of $\mathcal J_{\mathrm{DD}}$
and $\mathcal J_D$ in \eqref{eq:factor-graph-objective}.
With fixed model variances and unit robust weights, all three
configurations recover the same full-covariance Gaussian cost:
\begin{equation}
\begin{aligned}
    \left.
    \widetilde{\mathcal J}^{t}_{q,\mathrm B}
    \right|_{w^t_{q,\mathrm B}=1}
    =
    \left.
    \widetilde{\mathcal J}^{t}_{q,\mathrm S}
    \right|_{\mathbf W_q^t=\mathbf I} 
    =
    \min_{n_q^t}
    \left.
    \widetilde{\mathcal J}^{t}_{q,\mathrm L}
    \right|_{w_i^t=1\,\forall i} 
    =
    \frac12
    \|\mathbf r^t\|^2_{\boldsymbol\Sigma_q^t}.
\end{aligned}
\label{eq:robust-gaussian-equivalence}
\end{equation}

Thus, all three configurations retain the nominal shared-pivot
dependence; they differ in whether robust weights act on the
batch Mahalanobis norm, whitened components, or conditional
candidate residuals.

\subsection{Hybrid Ambiguity Continuation and IAR}
\label{subsec:ambiguity-management}

The carrier-phase integer ambiguity remains constant while tracking is
uninterrupted and no cycle slip occurs.
The phase residual in eqn.~\eqref{eq:dd-phase-residual} therefore requires the estimator to determine how ambiguity information is connected across epochs.
Two common ambiguity modes are continuous and instantaneous~\cite{wang2023give,teu2014inst}. 
Continuous modes either share one constant state
or connect successive states by random walks; instantaneous
mode uses independent per-epoch states
\cite{wang2023give,teu2014inst}.
Constant sharing gives stronger temporal coupling with fewer
variables but can propagate an undetected discontinuity;
random walks soften this coupling at the cost of additional
variables. Instantaneous mode discards temporal ambiguity
information.

SeA-RVINS uses a hybrid ambiguity-continuation strategy: verified continuous observations
share one constant ambiguity state over a bounded arc of up to $H$ consecutive GNSS epochs.
If continuity remains verified beyond the arc,
a new ambiguity state is introduced and connected to the preceding state by
a soft zero-change \ac{RW} factor.
When continuity is broken or cannot be verified, a new ambiguity state is
initialized independently. This preserves strong short-term ambiguity
coupling while limiting the duration of exact state sharing. We use $H=3$.

Cycle-slip detection is performed in \ac{GNSS} preprocessing using the
dual-frequency geometry-free test~\cite{blewitt1990automatic}.
Phase measurements for which no slip is detected are eligible for temporal ambiguity continuity. A detected slip, pivot change, or single-frequency measurement does not necessarily discard the \ac{DD} phase; instead, it prevents that measurement from inheriting a \ac{DD} ambiguity state from the preceding epoch~\cite{teu2014inst}.
For the bounded transition, the residual is
\begin{equation}
    r^N_{i,k}
    =\left(N_{\kappa_{i,k}}-N_{\kappa_{i,k-1}}\right)
    \sim\mathcal{N}(0,\sigma^2_{N,\mathrm{rw}}).
    \label{eq:ambiguity-random-walk}
\end{equation}

After the \ac{FGO} solves the current optimal estimate with float ambiguities,
the current-epoch ambiguity estimates
and their joint covariance are collected as $\widehat{\BoldN}$ and
$\BoldQ_{NN}$. The standard MLAMBDA algorithm~\cite{chang2005mlambda}
decorrelates these ambiguities and solves the integer least-squares problem
\begin{equation}
    \check{\BoldN}
    =\arg\min_{\mathbf n\in\mathbb Z^{m_A}}
      \|\widehat{\BoldN}-\mathbf n\|^2_{\BoldQ_{NN}}.
    \label{eq:integer-least-squares}
\end{equation}

Let $d_i$ denote the
conditional ambiguity variances obtained from the triangular decomposition of
the decorrelated covariance. The bootstrapped success rate~\cite{teunissen1998success} is computed as
\begin{equation}
    P_{\mathrm{B}}
    =\prod_{i=1}^{m_A}
      \left[
        2F_{\mathcal{N}}\!\left(\frac{1}{2\sqrt{d_i}}\right)-1
      \right],
    \label{eq:bootstrap-success-rate}
\end{equation}
where $F_{\mathcal{N}}(\cdot)$ is the standard-normal cumulative distribution
function.
The first-order position correction implied by the integer candidate is
\begin{equation}
    \delta\Boldp_{\mathrm{fix}}
    =\BoldQ_{pN}\BoldQ_{NN}^{-1}
      (\check{\BoldN}-\widehat{\BoldN}),
    \label{eq:ambiguity-position-innovation}
\end{equation}
where $\BoldQ_{pN}$ is the position--ambiguity cross covariance.
The candidate is accepted only if $P_{\mathrm B}>0.86$, the float pose
covariance passes its acceptance gate, and
$\delta\Boldp_{\mathrm{fix}}$ satisfies the designed
covariance and position innovation bounds. If validation fails, the current float estimate is retained without adding integer priors from that hypothesis.

Accepted \ac{IAR} fixed integers are imposed into \ac{FGO} through tight Gaussian ambiguity priors,
\begin{equation}
    r_i^{\mathrm{fix}}=N_{\kappa_i}-\check N_i,
    \qquad
    r_i^{\mathrm{fix}}\sim\mathcal N(0,\sigma^2_{\mathrm{fix}}),
    \label{eq:fixed-ambiguity-factor}
\end{equation}
The graph is reoptimized with the accepted ambiguity priors.
The \ac{RW} and fixing factors contribute to $\mathcal J_N$
in eqn.~\eqref{eq:factor-graph-objective}.
Shared ambiguity nodes impose
tighter geometric constraints on several subsequent epochs, and the resulting reduction in pose and ambiguity covariance also provides a more informative prediction for subsequent covariance-normalized measurement-outlier tests.




\section{Experiments}
\label{sec:experiments}

\subsection{Dataset and Evaluation Protocol}

Experiments use the public TEX-CUP dataset~\cite{narula2020tex}.
We exclude the initial 7 and final 9 minutes, when the vehicle is
parked in an open-sky lot, to prevent these open-sky intervals from dominating the statistics.
The evaluated route contains 4040 GNSS epochs and covers approximately 20~km through the west campus of
The UT Austin and downtown Austin.
It includes overpasses, high-rise buildings, large structures,
and dense foliage, with approximately 50\% deep-urban,
40\% light-urban, and 10\% open-sky driving.
It therefore evaluates both positioning accuracy and robustness
through prolonged GNSS-degraded conditions.
Representative street views are shown in Fig.~5 of~\cite{narula2020tex}.
Rover and base Septentrio receivers provide 1-Hz GPS L1/L2, Galileo E1/E5b,
and BeiDou B1I observations. The elevation cut-off is set at $15^\circ$.
A 100-Hz LORD MicroStrain IMU and synchronized 10-Hz Basler stereo images cropped to
$2048\!\times\!732$ pixels are used. Calibrated camera parameters and the antenna--IMU lever arm are applied. 
Ground truth trajectory data are available at GNSS epochs.
SeA-RVINS uses GTSAM~\cite{gtsam} for the \ac{FGO} solver with a 6-s fixed lag.
We use $c_{\mathrm{DCS}}=1.5$ for \ac{DCS} and a Huber
threshold of 1.345 throughout.

Table~\ref{tab:full_route_results} compares eight publicly
available baselines with the batch, scalar, and latent
configurations of SeA-RVINS. In the configuration columns,
$r$ and $b$ denote rover and base observations, respectively;
$r+b$ indicates that both are used. TC and LC denote tightly
and loosely coupled GNSS integration, respectively, while
M and S denote monocular and stereo vision. RTKLIB-EX~
\cite{rtklib_ex} serves as a standalone GNSS baseline and
provides the same position solutions to all LC methods.
The public methods are evaluated using the measurement
configurations they support. RTK-capable methods use rover
and base observations, whereas GVINS and InGVIO use
rover-only and are therefore included as system-level
references rather than controlled RTK-visual-inertial comparisons.
Necessary software bug fixes were applied to several
public implementations, including timing/clock handling
in GVINS, rotation normalization in VINS-Fusion, and state-index fix in IC-GVINS. GVINS and GICI-RRR also
required estimator adaptations.
RTKLIB-EX and GICI-RTK are used without source modifications.

\begin{figure}[t]
    \centering
    \includegraphics[width=0.9\columnwidth]{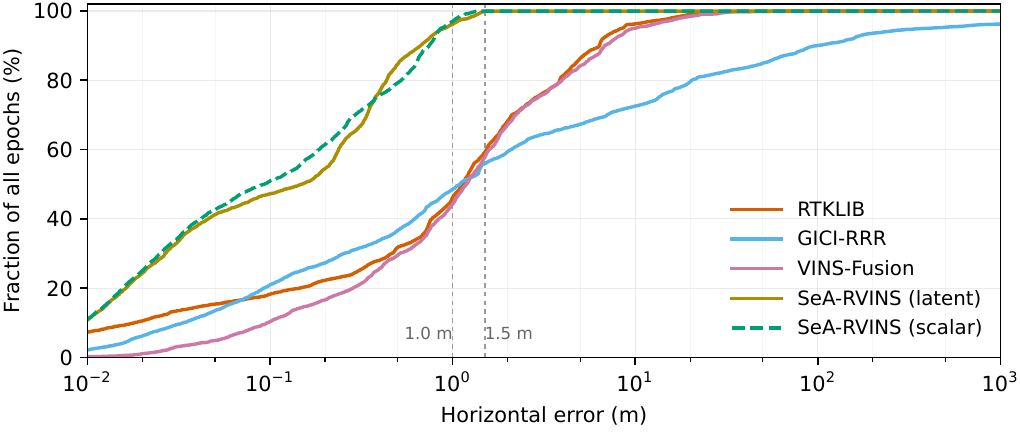}
    \caption{Cumulative horizontal-error distributions for selected
    methods. Vertical dashed lines mark 1.0 and 1.5~m.}
    \label{fig:error_cdf}
\end{figure}

All outputs are transformed to the GNSS antenna position and compared with ground truth at GNSS epochs over the same evaluation interval.
Horizontal and 3-D errors are the east--north norm and Euclidean position error, respectively. Availability is the fraction of evaluation epochs with a valid solution.
The percentages below 1.0 and 1.5~m and the curves
in Fig.~\ref{fig:error_cdf} are computed 
over all epochs, with missing solutions counted as unsuccessful.
\ac{RMSE}, maximum error, and 95th-percentile 
error (P95) are computed only over epochs with valid solutions.
In Table~\ref{tab:rtk_fix_results}, a used-fix epoch is one
at which an accepted integer solution is applied.
Fix rate is the fraction of evaluation epochs that are
used-fix epochs, and the associated horizontal-error
statistics are computed only over those epochs.

\begin{table*}[t]
    \centering
    \caption{Sensor configurations and positioning performance on the evaluated TEX-CUP route.}
    \label{tab:full_route_results}
    \scriptsize
    \setlength{\tabcolsep}{1.5pt}
    \renewcommand{\arraystretch}{1.10}
    \begin{tabular*}{\textwidth}{@{\extracolsep{\fill}}lccc*{9}{r}@{}}
        \toprule
        \multirow{2}{*}{\textbf{Method}} &
        \multicolumn{3}{c}{\textbf{Configuration}} &
        \multicolumn{5}{c}{\textbf{Horizontal error}} &
        \multicolumn{3}{c}{\textbf{3-D error}} &
        \multirow{2}{*}{\textbf{Avail. (\%)}} \\
        \cmidrule(lr){2-4}\cmidrule(lr){5-9}\cmidrule(lr){10-12}
        & \textbf{GNSS}
        & \textbf{Vision}
        & \textbf{IMU}
        & \mbox{RMSE (m)}
        & \mbox{Max. (m)}
        & \mbox{$<1.0$ m (\%)}
        & \mbox{$<1.5$ m (\%)}
        & \mbox{P95 (m)}
        & \mbox{RMSE (m)}
        & \mbox{Max. (m)}
        & \mbox{P95 (m)}
        & \\
        \midrule
        RTKLIB-EX & r+b & \ensuremath{\times} & \ensuremath{\times} & 4.95
        & 58.56 & 46.14\% & 59.23\% & 8.41 & 10.31
        & 66.99 & 21.52 & 100.0\% \\
        GICI-RTK~\cite{chi2023gici} & r+b & \ensuremath{\times} & \ensuremath{\times} & 4.69
        & 177.00 & 77.60\% & 87.43\% & 5.70 & 7.25
        & 265.26 & 8.64 & 96.7\% \\
        GICI-RRR~\cite{chi2023gici} & TC, r+b & M & \ensuremath{\checkmark} & 92.44
        & 1122.59 & 48.54\% & 55.97\% & 140.57 & 92.99
        & 1122.73 & 147.72 & 96.3\% \\
        VINS-Fusion~\cite{qin2025general} & LC & S & \ensuremath{\checkmark} & 5.10
        & 35.86 & 44.26\% & 57.82\% & 9.68 & 10.48
        & 59.07 & 22.38 & 99.9\% \\
        IC-GVINS~\cite{niu2022ic} & LC & M & \ensuremath{\checkmark} & 7.68
        & 30.47 & 9.16\% & 14.31\% & 21.95 & 8.10
        & 30.47 & 21.97 & 37.2\% \\
        GVINS~\cite{cao2022gvins} & TC, r & M & \ensuremath{\checkmark} & 436.70
        & $>\!1\mathrm{e}4$ & 22.48\% & 39.33\% & 28.76 & $>\!4\mathrm{e}4$
        & $>\!2\mathrm{e}5$ & 75.42 & 90.7\% \\
        InGVIO~\cite{liu2023ingvio} & TC, r & M & \ensuremath{\checkmark} & $>\!1\mathrm{e}3$
        & $>\!4\mathrm{e}4$ & 5.00\% & 6.58\% & 161.77 & $>\!1\mathrm{e}3$
        & $>\!4\mathrm{e}4$ & 223.53 & 98.4\% \\
        OKVIS2-X~\cite{boche2025okvis2} & LC & S & \ensuremath{\checkmark} & 19.29
        & 69.44 & 31.29\% & 44.98\% & 50.17 & 30.94
        & 91.05 & 73.07 & 100.0\% \\
        \midrule
        \textbf{SeA-RVINS (batch)} & TC, r+b & S & \ensuremath{\checkmark} & \textbf{0.38}
        & 2.91 & \textbf{97.15\%} & \textbf{99.63\%} & \textbf{0.90} & 1.30
        & 5.50 & \textbf{2.67} & 100.0\% \\
        \textbf{SeA-RVINS (scalar)} & TC, r+b & S & \ensuremath{\checkmark} & \textbf{0.39}
        & \textbf{1.63} & \textbf{97.00\%} & \textbf{99.98\%} & \textbf{0.86} & 1.73
        & 5.85 & 3.37 & 100.0\% \\
        \textbf{SeA-RVINS (latent)} & TC, r+b & S & \ensuremath{\checkmark} & \textbf{0.39}
        & \textbf{1.60} & \textbf{96.16\%} & \textbf{99.90\%} & \textbf{0.90} & \textbf{1.41}
        & \textbf{4.04} & 3.01 & 100.0\% \\
        \bottomrule
    \end{tabular*}
\end{table*}

\begin{table}[t]
    \centering
    \caption{\ac{IAR}-fixed solution summary. Horizontal-error statistics are conditional on used-fix epochs.}
    \label{tab:rtk_fix_results}
    \scriptsize
    \setlength{\tabcolsep}{1.2pt}
    \renewcommand{\arraystretch}{1.10}
    \begin{tabular*}{\columnwidth}{@{\extracolsep{\fill}}l*{6}{r}@{}}
        \toprule
        \multirow{2}{*}{\textbf{Method}} &
        \multirow{2}{*}{\shortstack{\textbf{Used-fix}\\\textbf{epochs}}} &
        \multirow{2}{*}{\shortstack{\textbf{Fix rate}\\\textbf{(\%)}}} &
        \multicolumn{4}{c}{\textbf{Horizontal error}} \\
        \cmidrule(lr){4-7}
        & & &
        \shortstack{$<0.5$ m\\(\%)} &
        \shortstack{$\geq 0.5$ m\\(epochs)} &
        \shortstack{RMSE\\(m)} &
        \shortstack{Max.\\(m)} \\
        \midrule
        SeA-RVINS (batch) & 1383 & 34.23 & 97.54 & 34 & 0.17 & 0.98 \\
        SeA-RVINS (scalar) & 1915 & 47.40 & 93.99 & 115 & 0.20 & 0.85 \\
        SeA-RVINS (latent) & 1842 & 45.59 & \textbf{99.35} & \textbf{12} & \textbf{0.15} & \textbf{0.55} \\
        \bottomrule
    \end{tabular*}
\end{table}

\begin{figure}[t]
    \centering
    \includegraphics[width=0.9\columnwidth]{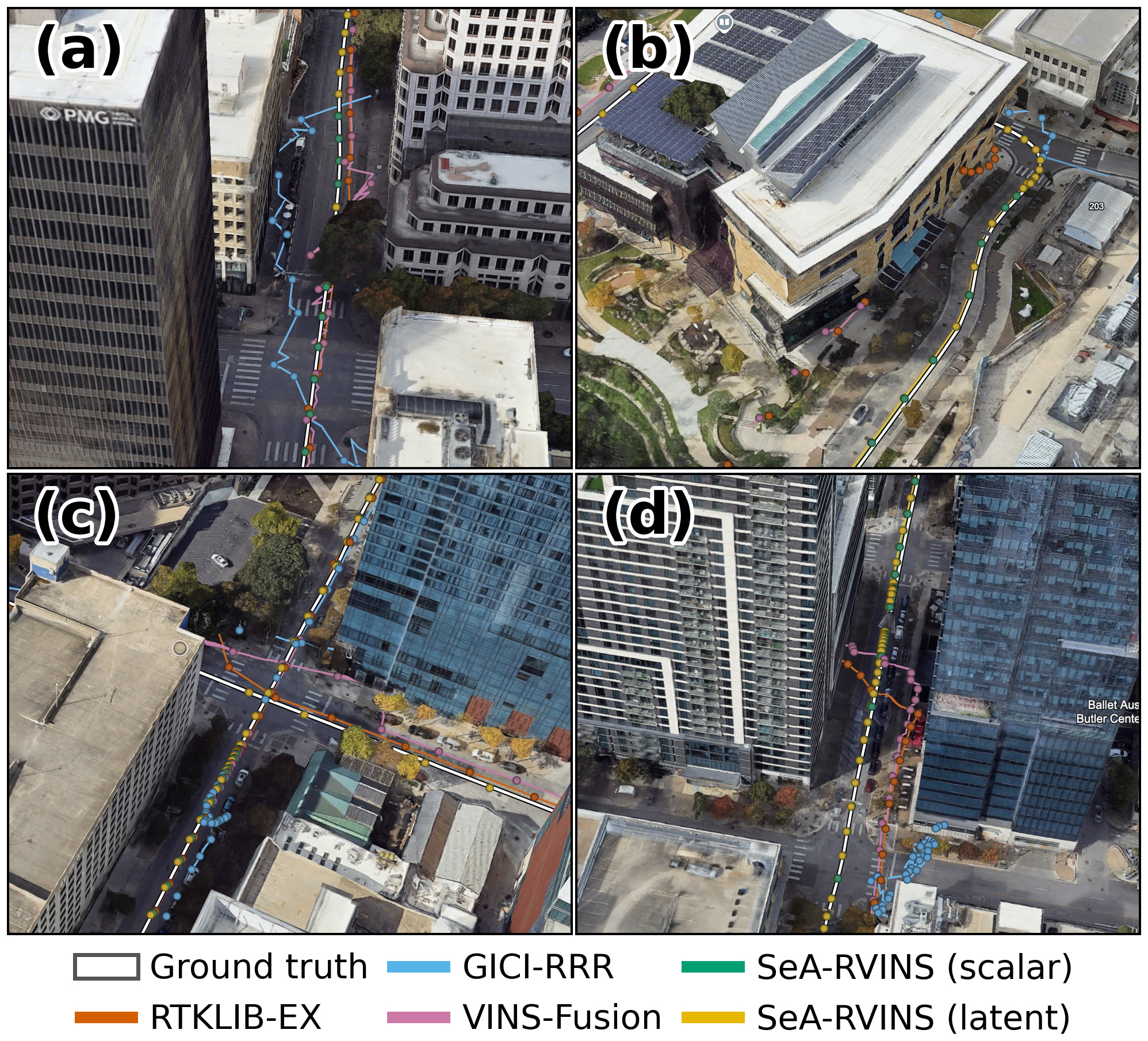}
    \caption{Estimated trajectories and ground truth in four selected
    downtown Austin segments. Imagery: \copyright~2026 Google
    (imagery date: Dec. 3, 2023).}
    \label{fig:3d_view}
\end{figure}

\subsection{Results and Discussion}

Table~\ref{tab:full_route_results} and
Fig.~\ref{fig:error_cdf} show that all three SeA-RVINS
configurations substantially outperform the evaluated public
baselines while maintaining 100\% availability. SeA-RVINS
achieves horizontal RMSEs of 0.38--0.39~m, with
96.16--97.15\% of all epochs below 1~m and maximum
horizontal errors of only 1.60--2.91~m.
In contrast, the evaluated public
baselines exhibit larger errors, divergence or reduced availability.
Because all LC baselines use the same RTKLIB-EX position
inputs, their results reflect
both upstream positioning quality and the downstream fusion
policy, not coupling architecture alone.
IC-GVINS diverges, yielding 37.2\% availability.
GICI-RRR also performs worse than GICI-RTK despite adding
visual--inertial measurements.
Although this is not a controlled frontend ablation, it shows that adding sensor measurements alone does not ensure robust estimation.
The SeA-RVINS results are consistent with the intended role
of its learned frontend: reliable landmarks provide relative
constraints under degraded GNSS, while semantic and
geometric screening reduce the risk of corrupted visual
factors entering the estimator. The low maximum errors and
100\% availability demonstrate robustness beyond average
positioning accuracy.

The three robust configurations achieve comparable
route-level accuracy but exhibit different characteristics.
Batch gives the lowest horizontal RMSE (0.38~m), scalar has
the lowest horizontal P95 (0.86~m), and latent has the lowest
maximum horizontal and 3-D errors (1.60 and 4.04~m).
The difference is clearer for \ac{IAR} in
Table~\ref{tab:rtk_fix_results}.
Scalar and latent increase used-fix rates from the batch configuration's 34.23\% to 47.40\% and 45.59\%, respectively.
Latent has 12 used-fix epochs with horizontal errors
of $\geq$ 0.5~m, compared with 34 for batch and 115 for
scalar, and the lowest RMSE and maximum error
(0.15 and 0.55~m).
In this evaluation, the latent-pivot configuration therefore accompany improved worst-case and used-fix accuracy, rather than uniformly better average accuracy.
The configurations therefore provide different
complexity--robustness trade-offs. Batch applies one weight
to the residual batch without additional graph
variables. Scalar assigns separate weights after
covariance whitening without additional variables.
Latent enables candidate-wise weighting by introducing
epoch-local pivot variables, and gives the best maximum error and fixed-solution accuracy in this evaluation.

Fig.~\ref{fig:3d_view} provides spatial examples of these
differences in downtown Austin. The SeA-RVINS trajectories
remain close to ground truth through the selected urban
corridors, whereas several baseline trajectories depart substantially,
including across building footprints. These examples
illustrate the large-error events in
Table~\ref{tab:full_route_results}.
\section{Conclusion and Future Work}
\label{sec:conclusion}

This paper presented SeA-RVINS, a fixed-lag
RTK--visual--inertial estimator for robust urban navigation.
It combines a semantic-aware learned stereo frontend,
hybrid ambiguity continuation, and correlation-preserving
robust GNSS estimation to limit the admission, influence,
and temporal propagation of unreliable measurements.
On the approximately 20-km TEX-CUP route, all three robust
configurations achieve 100\% availability and horizontal RMSEs
of 0.38--0.39~m. The latent-pivot configuration limits the
maximum horizontal error to 1.60~m and achieves the lowest
maximum 3-D error and the most accurate used-fix solutions.

Future work includes partial \ac{IAR} when the full ambiguity
vector cannot be reliably fixed, together with dedicated
robustness validation, and coarse lane-map aiding using lane centerlines and
boundaries as guarded lateral and heading constraints. The
latter will be particularly useful during long GNSS-blocked segments
in harsh urban environments, especially narrow streets where
satellite visibility is severely restricted.

\ificrafinal
  \section*{Acknowledgment}

The authors thank Yulin Xu (yulinxu@usc.edu) for his voluntary help in polishing Fig.~\ref{fig:system_overview} and Fig.~\ref{fig:vision_frontend}

\fi

\bibliographystyle{IEEEtran}
\bibliography{references}

\end{document}